%% file: ijcai20.tex
\documentclass{article}
\usepackage{ijcai20}

\usepackage{times}
\usepackage{soul}
\usepackage{url}
\usepackage[hidelinks]{hyperref}
\usepackage[utf8]{inputenc}
\usepackage[small]{caption}
\usepackage{graphicx}
\usepackage{amsmath}
\usepackage{amsthm}
\usepackage{booktabs}
\usepackage{algorithm}
\usepackage{algorithmic}
\usepackage{dingbat}
\usepackage{multirow}
\title{A Survey on Using Gaze Behaviour for Natural Language Processing}

\author{
Sandeep Mathias$^1$
\and
Diptesh Kanojia$^{1,2}$\and
Abhijit Mishra$^3$\And
Pushpak Bhattacharyya$^1$
\affiliations
$^1$Indian Institute of Technology Bombay\\
$^2$IITB-Monash Research Academy\\
$^3$Apple Inc., USA\\
\emails
\{sam, diptesh, pb\}@cse.iitb.ac.in,
abhijitmishra@apple.com
}

\begin{document}

\maketitle

\begin{abstract}
\input{Abstract.tex}
\end{abstract}

\input{Paper.tex}

\bibliographystyle{named}
\bibliography{ijcai2020}
\end{document}

%% file: Abstract.tex
Gaze behaviour has been used as a way to gather cognitive information for a number of years. In this paper, we discuss the use of gaze behaviour in solving different tasks in natural language processing (NLP). The collection of gaze behaviour is a costly task, both in terms of time and money. Hence, in this paper, we focus on research done to learn gaze behaviour in the models during run time. We also describe different eye tracking corpora in multiple languages. We conclude our paper by discussing applications in a domain - education - and how learning gaze behaviour can help in solving the tasks of complex word identification and automatic essay grading.

%% file: Paper.tex
\section{Introduction}
\label{Introduction Section}

Collecting psycholinguistic information from a reader has benefited multiple tasks in NLP, like named-entity recognition (NER) \cite{hollenstein-zhang-2019-entity}, text quality rating prediction \cite{mathias-etal-2018-eyes}, sarcasm understandability \cite{mishra2016predicting}, \textit{etc}. Gaze behaviour, in particular, has been shown to correlate well with cognitive processing of text, via the eye-mind hypothesis, which states that "there is no appreciable lag between what is fixated and what is processed" \cite{just1980theory}. Therefore, while using gaze behaviour is helpful for solving NLP tasks, a massive challenge involved is \textit{how do we collect the gaze behaviour in the first place}? In this paper, we describe research that uses gaze behaviour at run time to solve different NLP tasks.

While studying the gaze behaviour of a reader, we define the following terms. An \textbf{Interest Area} is the part of the screen that is of interest for analysis. In the case of NLP, it is mainly the words, although it could also be phrases, sentences, and paragraphs. A \textbf{Fixation} is an event where the eye is focused on the screen for a while. Fixations happen when the eye is processing the text. A \textbf{Saccade} is the movement of the eye from one fixation point to the next. There are 2 types of saccades - \textbf{Progressions} and \textbf{Regressions}. Progressions take place when there is a saccade from a fixation in the current interest area to a fixation in a later interest area. Regressions take place when there is a saccade from a fixation in the current interest area to a fixation in an earlier interest area.

\begin{figure*}[t]
\centering
\resizebox{\textwidth}{!}{
\includegraphics{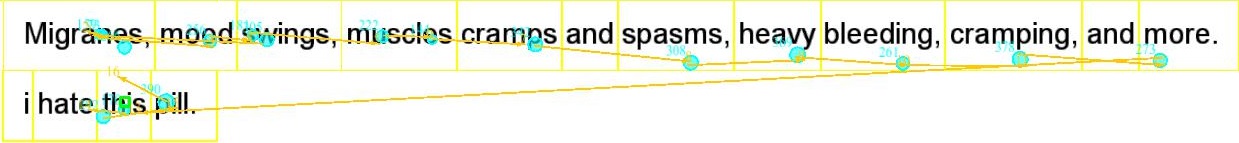}
}
\caption{Definitions of different terms used for gaze behaviour. The yellow boxes represent interest areas, the cyan circles represent fixations, and the dark yellow arrows correspond to saccades. Source: [Mishra \textit{et al.}, 2016c].}
\label{Definitions Figure}
\end{figure*}

Figure \ref{Definitions Figure} shows the output of an eye-tracker (the SR Research Eye Link 1000), for a reader reading a piece of text. The yellow boxes correspond to the interest areas, the cyan circles correspond to fixations, and the dark-yellow arrows correspond to saccades.

\section{Motivation for Learning Gaze Behaviour}
\label{Motivation Section}

The gaze behaviour of a reader can provide valuable psycholinguistic information for systems to help them solve tasks that require intelligence. For example, using gaze behaviour, we can better quantify the difficulty of sentences for translation \cite{mishra-etal-2013-translation-complexity-index}, or verify if a reader understands the sarcasm in a piece of text \cite{mishra2016predicting}, or even evaluate the quality of word embeddings \cite{sogaard-2016-evaluating}.

As recently as 2018, papers in the area of psycholinguistics, would describe work where mobile / portable eye-tracking systems would be available. \cite{mishra2016leveraging} mentioned that there would soon be an availability of eye-trackers on handheld devices like smartphones and tablets. \cite{mathias-etal-2018-eyes} also mentioned that SR Research have released a portable eye tracker. All these systems require a reader \textit{to read the text}, in order to collect gaze behaviour information. Our goal in this paper is to introduce the AI audience to ways in which we alleviate the need for readers to read texts at run time for the purpose of collecting gaze behaviour.


\section{Eye Tracking Corpora}
\label{Corpora Section}

There are several publicly available corpora where readers solve different tasks in different languages\footnote{A number of those corpora can be downloaded from \href{https://github.com/norahollenstein/cognitiveNLP-dataCollection/tree/master/eye-tracking}{here}.}. One of the earliest publicly available eye-tracking corpus is the Dundee Corpus \cite{kennedy2003dundee}. The Dundee Corpus was created for both English and French, with the English version having 20 articles from \textit{The Independent} read by 10 readers, and the French version having 20 articles from \textit{Le Monde} read by 10 French speakers.

There are a lot of corpora available in English. Some of them merely capture a reader's gaze behaviour as they read texts varying from articles to a novel\footnote{The novel read in the GECO Corpus is \textit{The Msterious Affair at Styles} by Agatha Christie.} \cite{kennedy2003dundee,hollenstein2018zuco,cop2017presenting,luke2018provo,yaneva2016assessing,mishra-etal-2017-scanpath}. Others capture the reader's gaze movement as the readers solve different NLP tasks \cite{mathias-etal-2018-eyes,cheri-etal-2016-leveraging,mishra2016predicting,mishra-EtAl:2016:P16-1,joshi-etal-2014-measuring}.

Along with English, there are a number of other publicly available corpora in other languages, like Chinese \cite{zang2018investigating,li2018understanding}, Dutch \cite{cop2017presenting,mak2019mental}, German \cite{nicenboim2016high,kliegl2004length} Persian \cite{safavi2016dependency}, Russian \cite{laurinavichyute2017russian}, and Spanish \cite{nicenboim2016high}. Table \ref{Eye Tracking corpora} gives the statistics for each of the publicly available eye-tracking corpora.

\begin{table*}[t]
\centering
\begin{tabular}{|l|l|c|l|l|}
\hline
\textbf{Dataset} & \textbf{Source} & \textbf{Language} & \textbf{Stimulus} & \textbf{Total Subjects} \\ \hline
Zang et al. (2018) & \cite{zang2018investigating} & \multirow{2}{*}{Chinese} & 90 sentences & 35 \\
Li et al. (2018) & \cite{li2018understanding} & & 15 documents & 29 \\ \hline
GECO & \cite{cop2017presenting} & \multirow{2}{*}{Dutch} & 1 novel & 33 \\
Mak \& Willems (2019) & \cite{mak2019mental} & & 3 short stories & 102 \\ \hline
ZuCo & \cite{hollenstein2018zuco} & \multirow{12}{*}{English} & 1107 sentences & 12 \\
GECO & \cite{cop2017presenting} & & 1 novel & 33 \\
Parker et. al (2017) & \cite{parker2017predictability} & & 40 documents & 48 \\
PROVO & \cite{luke2018provo} & & 55 texts & 84 \\
ASD Data & \cite{yaneva2016assessing} & & 27 documents & 27 \\
CFILT-Quality & \cite{mathias-etal-2018-eyes} & & 30 documents & 20 \\
CFILT-Scanpath & \cite{mishra-etal-2017-scanpath} & & 32 documents & 16 \\
CFILT-Coreference & \cite{cheri-etal-2016-leveraging} & & 22 documents & 14 \\
CFILT-Sarcasm & \cite{mishra2016predicting} & & 1000 tweets & 7 \\
CFILT-Sentiment & \cite{joshi-etal-2014-measuring} & & 1059 sentences & 5 \\
UCL Corpus & \cite{frank2013reading} & & 205 sentences & 43 \\
Dundee & \cite{kennedy2003dundee} & & 20 documents & 10 \\ \hline
Dundee & \cite{kennedy2003dundee} & French & 20 documents & 10 \\ \hline
Self-Paced Reading Time & \cite{nicenboim2016high} & \multirow{2}{*}{German} & 176 sentences & 72 \\
Potsdam Sentence Corpus & \cite{kliegl2004length} & & 144 sentences & 55 \\ \hline
Dependency Resolution dataset & \cite{safavi2016dependency} & Persian & 136 sentences & 40 \\ \hline
Russian Sentence Corpus & \cite{laurinavichyute2017russian} & Russian & 144 sentences & 96 \\ \hline
Self-Paced Reading Time & \cite{nicenboim2016high} & Spanish & 212 sentences & 79 \\ \hline
\end{tabular}
\caption{Summary of eye-tracking datasets available today. The novel read in the GECO dataset is \textit{The Mysterious Affair at Styles} by Agatha Christie (in both Dutch and English). Dataset is the name of the dataset. Source is the reference where the dataset was published. Language is the language of the data in the dataset. Stimulus is the amount of text data that the annotators read. Total subjects is the number of subjects who participated in the creation of the corpus.}
\label{Eye Tracking corpora}
\end{table*}

\section{Tasks Where Gaze Behaviour is Used}
\label{Tasks Section}

One of the earliest works that postulated the utility of gaze behaviour was a study on reading comprehension done by Just and Carpenter \shortcite{just1980theory}. They came up with the eye-mind hypothesis, which stated that ``there is no appreciable lag between what is fixated by the eye and what is processed by the mind.'' Many studies, done in the field of cognitive science, have found relationships between different aspects of gaze behaviour and the corresponding aspects of text, such as relationships between fixations and word length \cite{rayner1998eye,henderson1993eye}, word predictability \cite{rayner1998eye}, etc.

Gaze behaviour is also beneficial for multiple tasks in NLP where the reader's psychological input is critical, and text features alone will not be enough. \cite{mishra-etal-2013-translation-complexity-index} discuss how using gaze behaviour would be a better approach to judge the complexity of translating a sentence, compared to using just length-based statistics (like word length, sentence length, \textit{etc.}). Understanding sarcastic texts could also be resolved using gaze behaviour, where incongruity in the text (one of the leading indicators of sarcasm) induces gaze behaviour characterized by longer fixations, regressions \textit{etc.} \cite{mishra2016predicting}. Gaze behaviour has also been used to identify a reader's native language \cite{berzak-etal-2017-predicting}, as well as in detecting grammatical errors in compressed sentences \cite{klerke-etal-2015-looking}. \cite{klerke2015reading} also show that gaze behaviour can be used to evaluate the output of Machine Translation systems better than automated metrics. Gaze behaviour has also been used to evaluate how a reader would rate the quality of a piece of text \cite{mathias-etal-2018-eyes}.

The scanpath of a reader (\textit{i.e.} the path a reader's eye traverses when they read the text) has been used to test how easy/difficult a piece of text is for a reader to read \cite{mishra-etal-2017-scanpath}. It can also be used to predict the misreadings among children with reading difficulties \cite{bingel2018predicting}.

Gaze behaviour has also been used in multiple areas of sentiment analysis such as sarcasm detection \cite{mishra-EtAl:2016:P16-1}, sarcasm understanding \cite{mishra2016predicting}, and sentiment analysis annotation tasks \cite{joshi-etal-2014-measuring}. \cite{klerke2019glance} tackle the classical problem of POS (Part-of-speech) tagging with gaze data and provide a systematic overview of the influence of two independent levels of gaze data aggregation on low-level syntactic labelling tasks at two separate levels of complexity; \textit{i.e.,} a simple chunk boundary tagging and a supervised POS tagging task.

\section{Learning Gaze Behaviour}
\label{Gaze Behaviour Learning Section}
As mentioned in the previous section, using gaze behaviour helps systems in solving many NLP tasks. However, \textit{learning} gaze behaviour is still a relatively new challenge. The biggest benefit of learning gaze behaviour at run time is that we alleviate the requirement of reading the texts in order to get gaze behaviour information. In this section, we look at different tasks and different systems which alleviate the need for collecting gaze behaviour at run time.

\subsection{NLP Tasks}

\paragraph{Reading} \cite{nilsson-nivre-2009-learning} describe an approach to detect which tokens readers fixate on while reading, using a transition-based approach. \cite{matthies-sogaard-2013-blinkers} improve on their approach using a linear CRF model. While fixation accuracy and F1 are comparable to \cite{nilsson-nivre-2009-learning}'s approach, \cite{matthies-sogaard-2013-blinkers}'s approach predicts the eye movements better for new readers, than when training and test data come from the same reader.

\paragraph{Text Simplification} \cite{klerke-etal-2016-improving} describe a way to simplify text by compressing sentences using gaze behaviour learnt at run time. They used the Ziff-Davis \cite{knight2002summarization}, Broadcast \cite{clarke-lapata-2006-models}, and Google \cite{filippova-etal-2015-sentence} datasets.

\paragraph{Part-of-Speech (PoS) tagging} \cite{barrett-etal-2016-pos-tagging} describe an approach to solve the task of part-of-speech tagging using gaze behaviour from the Dundee Treebank \cite{barrett2015dundee}. \cite{barrett-etal-2016-cross} investigate the same using cross-lingual approaches (i.e. training on English, test on French, and vice versa).

\paragraph{Readability} \cite{gonzalez-garduno-sogaard-2017-gaze-readability} describe a solution for predicting readability using multi-task learning, with readability prediction as the primary task, and predicting gaze behaviour as an auxiliary task.

\paragraph{Sentiment Analysis} \cite{mishra2018cognition} describe a way to use gaze behaviour for the task of sentiment analysis of movie reviews. They use a multi-task framework, where the primary task is predicting the sentiment, and the auxiliary tasks are part-of-speech tagging and learning gaze behaviour. They showed statistically significant improvements over the state-of-the-art reported results using both their multi-task systems on the IMDB25K dataset \cite{maas-etal-2011-learning}, and the PL2000 dataset \cite{pang-lee-2004-sentimental}.

\paragraph{NER} \cite{hollenstein-zhang-2019-entity} describe a way to learn gaze behaviour at run time, and show improvements on the CoNLL 2003 dataset \cite{tjong-kim-sang-de-meulder-2003-introduction} by learning gaze behaviour at run time.

\paragraph{Multiple NLP Tasks} \cite{barrett-etal-2018-sequence} describe a solution to multiple NLP tasks - sentiment analysis, grammatical error detection, and hate speech detection - using multi-task learning. For each sentence, they learn gaze features at the word-level and solve the NLP task at the sentence level.

\subsection{System Architectures}

\begin{table*}[t]
\centering
\resizebox{\textwidth}{!}{%
\begin{tabular}{|l|l|l|c|c|c|c|}
\hline
\textbf{NLP Task} & \textbf{Dataset} & \textbf{Gaze System} & \textbf{Best Baseline} & \textbf{Gaze Result} & \textbf{\%age Increase} & \textbf{SS?} \\ \hline
Reading & Dundee Corpus & \cite{matthies-sogaard-2013-blinkers} & 57.70 & 69.90 & 12.2 & Yes \\
Text Simplification & Google & \cite{klerke-etal-2016-improving} & 79.80 & 80.97 & 1.17 & N/R \\
PoS Tagging & Dundee Corpus & \cite{barrett-etal-2016-pos-tagging} & 79.77 & 82.44 & 2.67 & Yes \\
Readability & Wikipedia & \cite{gonzalez-garduno-sogaard-2017-gaze-readability} & 78.87 & 86.45 & 9.61 & Yes \\
Sentiment Analysis & Twitter & \cite{barrett-etal-2018-sequence} & 49.67 & 52.60 & 2.93 & Yes \\
Hate Speech Detection & \cite{waseem-hovy-2016-hateful} & \cite{barrett-etal-2018-sequence} & 74.16 & 75.61 & 1.45 & Yes \\
NER & CoNLL-2003 & \cite{hollenstein-zhang-2019-entity} & 63.92 & 66.61 & 4.21 & Yes \\
\hline
\end{tabular}%
}
\caption{Sample results for different NLP tasks using gaze behaviour learnt at test time. Dataset is the name of one of the datasets that was used. Best Baseline corresponds to the result of the best baseline system, gaze result corresponds to the result of the gaze behaviour system on that dataset, \%age increase is the percentage improvement for that NLP task, and $SS?$ says whether or not the improvements were statistically significant. Due to space constraints we are only showing the results for 1 dataset per task. N/R means that the authors haven't reported the results of any statistical significance test.}
\label{Sample Results Table}
\end{table*}

\cite{reichle2003ez} describe a theoretical framework (E-Z Reader) for understanding how word identification, visual processing, attention, and oculomotor control jointly determine when and where the eyes move during reading.

\cite{engbert2005swift} propose a mathematical model for the control of eye movements during reading that is both psychologically and neurophysiologically plausible and that accounts for most of the known experimental findings. This model is an extension of the model they had proposed previously \cite{engbert2002dynamical}.

\cite{nilsson-nivre-2009-learning} use a transition-based model of reader's eye movements to predict the next word that is fixated. They used features like token length, token frequency class, next token length, next token frequency class, etc. \cite{matthies-sogaard-2013-blinkers} use a linear CRF to determine which words are fixated on while reading. The features that they use are word length (for a window of 5 words), and word probability (for a window of 3 words).

\cite{klerke-etal-2016-improving} describe two multi-task learning models to use gaze behaviour. The first (Multi-task) uses multi-task learning with a separate logistic regression classifier for predicting gaze behaviour during training. The second (Cascaded) uses multi-task learning with gaze behaviour predicted in an inner layer.

\cite{barrett-etal-2016-pos-tagging} use a Hidden Markov Model with additional type-aggregated gaze features, like first fixation duration, fixation probabilities of neighbouring words, etc. They used a series of features classified as Early (first fixation duration, previous word fixation probability, etc.), Late (total regression-to duration, re-read probability, etc.), Basic (total fixation duration, mean fixation duration, etc.), Regression From (regressions from a word, long regressions from a word, etc.), Context (fixation probability and duration of nearby words), NoGaze features (word length, probability in the BNC and Dundee corpora, etc.).

\cite{gonzalez-garduno-sogaard-2017-gaze-readability} use a multi-task multi-layer perceptron and multi-task logistic regression systems, with their best result coming from using all features using the multi-task multi-layer perceptron. The gaze behaviour is learnt using the Dundee Corpus \cite{kennedy2003dundee}.

\cite{mishra2018cognition} use a multi-task learning approach to learn gaze behaviour and perform PoS tagging as auxiliary tasks while predicting the sentiment of the review as the primary task. They use a pair of Bi-Directional LSTMs, in which one bi-LSTM learns the first fixation duration, while the other performs PoS tagging.

\cite{hollenstein-zhang-2019-entity} learn gaze behaviour from multiple corpora - the Dundee Corpus \cite{kennedy2003dundee}, the GECO Corpus \cite{cop2017presenting}, and the ZuCo Corpus \cite{hollenstein2018zuco} - and use a bi-directional LSTM neural architecture with conditional random fields to perform NER. For each word, their system takes character embeddings, word embeddings and several gaze behaviour features as input, and outputs the corresponding NER tag.

\subsection{Normalizing Data}
While collecting gaze behaviour data, it is better to normalize gaze behaviour across readers. Gaze behaviour is normalized in the following ways. \textbf{Min-Max Normalization} is where we normalize the data on a scale of $[0,1]$, where $0$ corresponds to the lowest value, and $1$ corresponds to the highest value. \cite{barrett-etal-2016-pos-tagging} is one work that uses min-max normalization. \textbf{Binning} is where we consider discrete bins for each task. For example: 6 bins (numbered 0 to 5) as described by \cite{klerke-etal-2016-improving}. One of the advantages of binning over min-max normalization is that it reduces the effect of outliers in the data.

\subsection{Results}

Table \ref{Sample Results Table} gives the results of each of the different NLP tasks where we don't record gaze behaviour at run time. Due to space constraints, we report the result of only 1 result and dataset for each NLP task. We report the best baseline result, the corresponding gaze behaviour system result, the improvement in percentage points, and whether or not the improvement in performance was statistically significant.

\section{Applications of Learning Gaze Behaviour}
\label{Applications Section}
In this section, we cover some applications in NLP which we believe could benefit a lot from \textit{learning} gaze behaviour. To the best of our knowledge, there is no work on any of these applications where the system learns gaze behaviour.

\subsection{Complex Word Identification}

Lexical simplification is a process in which text gets simplified by replacing complex words and phrases with simpler ones. For example, a non-native speaker of English will struggle to understand what the word ``vituperate'' means, in the sentence ``The Nazi propaganda vituperated the Jews.''\footnote{\href{http://wordnetweb.princeton.edu/perl/webwn?s=vituperate&sub=Search+WordNet&o2=&o0=1&o8=1&o1=1&o7=&o5=&o9=&o6=&o3=&o4=&h=111302201113000000000000022200}{WordNet entry for vituperate}.} However, they will be more likely to understand the meaning of the sentence ``The Nazi propaganda vilified the Jews''. The process of identifying which words are hard (and should be replaced by an appropriate synonym) is a very useful application for the use of eye-tracking, as readers may fixate longer on harder words \cite{rayner1998eye}.

\cite{paetzold-specia-2016-semeval} report the results of the shared task on complex word identification held at Sem-Eval 2016. Another shared task was organized in 2018, to identify complex words, and phrases, in English, Spanish, German, and a cross-lingual setting (where the target language was French) \cite{yimam-etal-2018-report}.

Using cognitive information should aid in solving these tasks. As mentioned earlier, there are quite a few papers that deal with text complexity / simplification and readability which use gaze behaviour like  \cite{klerke-etal-2016-improving}, \cite{mishra-etal-2017-scanpath}, \cite{gonzalez-garduno-sogaard-2017-gaze-readability}, \cite{gonzalez2018learning}, etc. However, in proposing solutions for the shared tasks, none of them used cognitive information. With the availability of a large number of eye-tracking corpora, in various languages, an interesting avenue of research would be exploring how using gaze behaviour can help in identifying complex words, even if we don't have such information at run time.

\subsection{Automatic Essay Grading}

An essay is a text, written in response to a topic, called the essay prompt. Grading an essay is assigning a score to the essay based on its quality, either for the essay on the whole (holistic scoring) or for certain aspects of the essay (trait-specific scoring) \cite{ijcai2019-879}. Automatic essay grading (AEG) is the process of grading an essay using a machine. The earliest AEG system was described by \cite{page1966imminence}, over 50 years ago. Since then, there have been a number of commercial AEG systems, like E-Rater \cite{attali2006automated}, Intelligent Essay Assessor \cite{landauer2003automated}, LightSide \cite{mayfield2013lightside}, \textit{etc}.

The current state-of-the-art AEG systems use neural networks, like CNNs \cite{dong-zhang-2016-automatic}, LSTMs \cite{alikaniotis-yannakoudakis-rei:2016:P16-1,taghipour-ng-2016-neural,tay-2018-skipflow}, or both \cite{dong-etal-2017-attention,zhang-litman-2018-co}. The dataset that they use is the 2012 Automatic Student Assessment Prize (ASAP) AEG Dataset, released by the Hewlett Foundation \cite{ijcai2019-879}.

As mentioned earlier, \cite{mathias-etal-2018-eyes} describe a way to predict the rating a reader would give a piece of text based on its quality. Their work showed that gaze behaviour we could use for similar applications like AEG. However, \cite{mathias-etal-2018-eyes}'s approach required readers to read the text in order to use the reader's gaze behaviour. An AEG system built using multi-task learning where the primary task is scoring the essay at the document-level, and learning the gaze behaviour is the auxiliary task at the word-level.  \cite{barrett-etal-2018-sequence} have shown this approach to benefit multiple NLP tasks like sentiment analysis, grammatical error detection and hate speech detection.

\section{Conclusion}
\label{Conclusion Section}


Gaze behaviour has been shown to aid multiple natural language processing tasks \cite{mishra2018cognitively}. However, collecting gaze behaviour at run time is not feasible. Hence, in order to use gaze behaviour, we utilize different approaches, like multi-task learning, using type-aggregated values, etc.

In this paper, we first introduce the AI audience to different NLP tasks which are solved using gaze behaviour, like translation complexity, sarcasm understandability, text quality rating prediction, etc. We then discuss different tasks where we show that gaze behaviour aids in their solution.

To solve any of these tasks, we require gaze behaviour data to be ready for training. In our paper, we also report gaze behaviour datasets created in multiple languages. Finally, we describe a pair of applications from the domain of education - complex word identification, and automatic essay grading - which could benefit a lot from using gaze behaviour based solutions.